# Task-Aware Compressed Sensing with Generative Adversarial Networks

**Maya Kabkab**\*, **Pouya Samangouei**\*, and **Rama Chellappa**
Department of Electrical and Computer Engineering
University of Maryland Institute for Advanced Computer Studies
University of Maryland, College Park, MD 20742
{mayak, pouya, rama}@umiacs.umd.edu

## Abstract

In recent years, neural network approaches have been widely adopted for machine learning tasks, with applications in computer vision. More recently, unsupervised generative models based on neural networks have been successfully applied to model data distributions via low-dimensional latent spaces. In this paper, we use Generative Adversarial Networks (GANs) to impose structure in compressed sensing problems, replacing the usual sparsity constraint. We propose to train the GANs in a task-aware fashion, specifically for reconstruction tasks. We also show that it is possible to train our model without using any (or much) non-compressed data. Finally, we show that the latent space of the GAN carries discriminative information and can further be regularized to generate input features for general inference tasks. We demonstrate the effectiveness of our method on a variety of reconstruction and classification problems.

## Introduction

The broad scope and generality of the field of compressed sensing has led to many impressive applications, such as rapid magnetic resonance imaging [Lustig, Donoho, and Pauly, 2007], single-pixel camera [Duarte et al., 2008] and UAV systems. The core problem of compressed sensing is that of efficiently reconstructing a signal $\mathbf{x} \in \mathbb{R}^n$ from an under-determined linear system of noisy measurements given by

$$\mathbf{y} = \mathbf{A}\mathbf{x} + \boldsymbol{\zeta} \qquad (1)$$

where $\mathbf{A} \in \mathbb{R}^{m \times n}$ is the measurement sensing matrix, $m < n$, and $\boldsymbol{\zeta} \in \mathbb{R}^m$ is the measurement noise [Donoho, 2006]. Since this is an under-determined system of equations, a unique solution does not exist, even in the absence of noise, unless some assumptions are made on the structure of the unknown vector $\mathbf{x}$. Depending on applications, the structural assumptions may vary, the most common one being that $\mathbf{x}$ is sparse [Donoho, 2006, Candes, Romberg, and Tao, 2006b, Candes and Wakin, 2008]. Under this specific assumption, the problem of recovering $\mathbf{x}$ has been widely studied, and different conditions on the matrix $\mathbf{A}$ have been established to guarantee reliable recovery [Bickel, Ritov, and Tsybakov, 2009]. These conditions include the Restricted Isometry Property (RIP) or the Restricted Eigenvalue Condition (REC) [Candes, Romberg, and Tao, 2006a, Pati, Rezaiifar, and Krishnaprasad, 1993].

Even though the sparsity assumption on $\mathbf{x}$ is the most common choice, it is not the only possible one. Indeed, other approaches, such as combining sparsity with additional model-based constraints [Baraniuk et al., 2010] or graph structures [Hegde, Indyk, and Schmidt, 2015], have been developed. Recently, in [Bora et al., 2017], a *generative model* was used and the unknown signal $\mathbf{x}$ was assumed to be the output of this model. Generative models have been successfully used to model data distributions, and include the variational auto-encoder (VAE) [Kingma and Welling, 2013], generative adversarial networks (GANs) [Goodfellow et al., 2014], and variations thereof [Radford, Metz, and Chintala, 2016, Chen et al., 2016, Lamb, Dumoulin, and Courville, 2016]. In the GAN framework, two models are trained simultaneously in an adversarial setting: a generative model that emulates the data distribution, and a discriminative model that predicts whether a certain input came from real data or was artificially created. The generative model learns a mapping $G$ from a low-dimensional vector $\mathbf{z} \in \mathbb{R}^k$ to the high dimensional space $\mathbb{R}^n$.

The authors in [Bora et al., 2017] use a pre-trained generative model $G$, and recover an estimate of $\mathbf{x}$ from the compressed measurements $\mathbf{y}$, assuming it is in the range of $G$. To this end, the following optimization problem is solved:

$$\begin{aligned} \min_{\hat{\mathbf{x}}, \mathbf{z}} \quad & ||\mathbf{A}\hat{\mathbf{x}} - \mathbf{y}||_2^2 \\ \text{s.t.} \quad & \hat{\mathbf{x}} = G(\mathbf{z}) \end{aligned} \qquad (2)$$

In the setting of [Bora et al., 2017], the pre-trained $G$ is unaware of the compressed sensing framework. Furthermore, it is assumed that an abundance of real (non-compressed) images is available to train $G$, which, depending on the application, may not be a realistic assumption [Lustig et al., 2008]. After all, the aim of compressed sensing is to recover signals through the use of compressed measurements. In this paper, we propose to train $G$ specifically for the task of recovering compressed measurements. This makes our GAN task-aware, and improves the compressed sensing performance. Our approach will also address the case where no or very little non-compressed

---

data is available for training, by complementing the training set with compressed training data. Finally, we empirically demonstrate that the low-dimensional latent vector **z** can be used, not only to perform reconstruction via $G$, but also for inference tasks such as classification. Our code has been made publicly available at https://github.com/po0ya/csgan.

**Our contributions:**

1. We train the GAN in a task-aware fashion allowing it to be specifically optimized for the reconstruction task. We show that this consistently improves the reconstruction error obtained in [Bora et al., 2017] for various values of the number of measurements $m$.

2. We consider training using a combination of a small number of (or no) non-compressed data and a larger set of compressed training data. This is achieved by introducing a second discriminator specifically for compressed data.

3. We show that we can regularize the latent space of **z** to make it discriminative, given a desired inference task.

## Related Work

In this work, we combine compressed sensing and generative models to perform reconstruction and classification tasks. To this end, we explain the related work in two parts. The first part addresses the use of generative models for reconstruction and classification tasks, and the second part reviews inference tasks in compressed sensing.

Using a generative model for reconstruction tasks is a fairly well-researched area. One line of work attempts to map an image to the range of the generator [Dumoulin et al., 2016, Donahue, Krähenbühl, and Darrell, 2016, Lipton and Tripathi, 2017]. Unlike our setting, complete and non-compressed knowledge of the images is assumed. In [Lipton and Tripathi, 2017], gradient descent (GD) is used to project the image samples onto the latent space of a pre-trained generative model. In [Dumoulin et al., 2016, Donahue, Krähenbühl, and Darrell, 2016], an inverse mapping between the input space of **x** and the latent space of **z** is jointly learned along with the generator in an adversarial setting. Generative models can also be used for classification tasks [Mirza and Osindero, 2014, Springenberg, 2015, Odena, 2016, Lamb, Dumoulin, and Courville, 2016]. This can be achieved by modifying the discriminator of the GAN to also output class probabilities [Springenberg, 2015, Odena, 2016] or augmenting the loss function with discriminative features at training time [Mirza and Osindero, 2014, Lamb, Dumoulin, and Courville, 2016]. Such discriminative features include ground-truth class labels [Mirza and Osindero, 2014] and representations learned by a pre-trained classifier [Lamb, Dumoulin, and Courville, 2016].

Another related line of work considers compressed sensing frameworks for various machine learning and computer vision tasks [Davenport et al., 2007, Cevher et al., 2008, Maillard and Munos, 2009, Calderbank, Jafarpour, and Schapire, 2009, Lohit et al., 2015]. In [Calderbank, Jafarpour, and Schapire, 2009] theoretical results are provided showing that inference can be done directly in the compressed domain. Of particular relevance to our work are [Davenport et al., 2007, Lohit et al., 2015, Lohit, Kulkarni, and Turaga, 2016] which develop various techniques for the classification of compressed images. These methods operate directly on the compressed measurements, whereas we perform classification on the latent variable **z**.

Finally, one last research area that is relevant to our application is super-resolution, the task of increasing the resolution of an image. This can be seen as a special case of compressed sensing where the sensing matrix **A** averages neighboring pixels. In [Yang et al., 2010], a sparse representation of image patches is sought and used to obtain a high-resolution output. Our framework adopts the generative model instead of the sparsity constraint. More recent work uses deep convolutional networks [Dong et al., 2016, Kim, Kwon Lee, and Mu Lee, 2016].

## Model Description

### Background Information

Before describing our approach, we provide some necessary background information on compressed sensing and GANs.

In compressed sensing, the measurements are given as $\mathbf{y} = \mathbf{A}\mathbf{x} + \boldsymbol{\zeta}$. $\mathbf{A} \in \mathbb{R}^{m \times n}$ is the measurement matrix and is usually chosen to be a Gaussian random matrix because it satisfies desirable properties with high probability [Donoho, 2006]. Unless otherwise specified, we will assume that **A** is a zero-mean random Gaussian matrix with independent and identically distributed entries. **A** is kept constant in a given experiment.

GANs consist of two neural networks, $G$ and $D$. $G : \mathbb{R}^k \to \mathbb{R}^n$ maps a low-dimensional latent space to the high dimensional sample space of **x**. $D$ is a binary neural network classifier. In the training phase, $G$ and $D$ are typically learned in an adversarial fashion using actual input data samples **x** and random vectors **z**. An isotropic Gaussian prior is usually assumed on **z**. While $G$ learns to generate outputs $G(\mathbf{z})$ that have a distribution similar to that of **x**, $D$ learns to discriminate between "real" samples **x** and "fake" samples $G(\mathbf{z})$. $D$ and $G$ are trained in an alternating fashion to minimize the following min-max loss [Goodfellow et al., 2014]:

$$\min_G \max_D V(D,G) = \mathbb{E}_{\mathbf{x} \sim p_{\text{data}}(\mathbf{x})}[\log D(\mathbf{x})]$$
$$+ \mathbb{E}_{\mathbf{z} \sim p_{\mathbf{z}}(\mathbf{z})}[\log(1 - D(G(\mathbf{z})))] \quad (3)$$

**GAN Training Algorithm** At every iteration, (3) is maximized over $D$ for a fixed $G$, using GD, and then minimized over $G$, fixing $D$.

### Motivation

In [Bora et al., 2017], a generative model is pre-trained on a set of uncompressed training images, using the algorithm described in [Radford, Metz, and Chintala, 2016]. In the testing phase, the generative model is used to reconstruct a compressed, previously unseen, test image using GD on the problem in (2). It is shown that, when **A** is a random Gaussian matrix, if $\hat{\mathbf{z}}$ minimizes $||\mathbf{A}G(\mathbf{z}) - \mathbf{y}||_2$ to within additive $\epsilon$ of the optimum, then for all **x**, and with high probability

$$||G(\hat{\mathbf{z}}) - \mathbf{x}||_2 \leq 6 \min_{\mathbf{z}} ||G(\mathbf{z}) - \mathbf{x}||_2 + 3||\boldsymbol{\zeta}||_2 + 2\epsilon \quad (4)$$

In other words, the observed reconstruction error is bounded by the minimum possible error of any vector in the range of the generator with some additional terms due to noise and GD precision. We note that this upper bound depends on how well $G$ can represent the unknown signal $\mathbf{x}$. Now, we show that, under certain conditions, the expected value of this error term converges to 0 as $G$ is trained on $\mathbf{x}$.

**Theorem 1** *Let $G_t$ be the generator of a GAN after $t$ steps of the GAN training algorithm described above. Additionally, as in [Goodfellow et al., 2014], we assume:*

*(i) $G$ and $D$ have enough capacity to represent the data.*

*(ii) At each update, $D$ reaches its optimum given $G$.*

*(iii) At each update, $G$ is updated to improve the min-max loss in (3).*

*Furthermore, we assume that the training samples $\mathbf{x}$ come from a continuous distribution with compact support. Then,*

$$\lim_{t\to\infty} \mathbb{E}_{\mathbf{x}}\left[\min_{\mathbf{z}} ||G_t(\mathbf{z}) - \mathbf{x}||_2\right] = 0 \quad (5)$$

**Proof:** Let $g_t(\mathbf{x})$ be the probability distribution function (pdf) of $G_t(\mathbf{z})$ and $f(\mathbf{x})$ be the pdf of $\mathbf{x}$. Then, from [Goodfellow et al., 2014, Proposition 2], $g_t(\mathbf{x})$ converges to $f(\mathbf{x})$ pointwise in $\mathbf{x}$. By assumption, $f(\mathbf{x})$ has bounded support $\mathcal{X}$, i.e., $\mu(\mathcal{X})$ is finite, where $\mu(\cdot)$ is the Lebesgue measure. We note that the assumption of $\mathcal{X}$ having bounded support is reasonable, especially for computer vision tasks where pixel values are usually bounded (for instance, in $[0, 255]$).

Then, by Egorov's theorem, for all $\epsilon > 0$, there exists a set $B \subseteq \mathcal{X}$ such that $\mu(B) < \epsilon$ and $g_t(\mathbf{x})$ converges to $f(\mathbf{x})$ uniformly on $\mathcal{X} \setminus B$. This implies that, for all $\mathbf{x} \in \mathcal{X} \setminus B$ and for all $\nu$, there exists $t_0$ such that $|g_t(\mathbf{x}) - f(\mathbf{x})| < \nu$, for all $t \geq t_0$. This means that, for $\mathbf{x} \in \mathcal{X} \setminus B$, $g_t(\mathbf{x}) = 0$ implies $f(\mathbf{x}) < \nu$. Additionally, $g_t(\mathbf{x}) > 0$ implies that there exists $\mathbf{z}$ such that $G_t(\mathbf{z}) = \mathbf{x}$, i.e., $\min_{\mathbf{z}} ||\mathbf{x} - G_t(\mathbf{z})||_2 = 0$.

Let $\mathcal{X}_\nu = \{\mathbf{x} \in \mathcal{X} \mid f(\mathbf{x}) \leq \nu\}$. Note that $\{\mathbf{x} \in \mathcal{X} \setminus B \mid g_t(\mathbf{x}) = 0\} \subseteq \mathcal{X}_\nu$ for all $t \geq t_0$. Then, for all $\epsilon, \nu > 0$ and $t \geq t_0$,

$$\mathbb{E}_{\mathbf{x}}\left[\min_{\mathbf{z}} ||\mathbf{x} - G_t(\mathbf{z})||_2\right] \quad (6)$$

$$\leq \int_B \min_{\mathbf{z}} ||\mathbf{x} - G_t(\mathbf{z})||_2 \, f(\mathbf{x}) \, d\mathbf{x}$$

$$+ \int_{\mathcal{X}_\nu} \min_{\mathbf{z}} ||\mathbf{x} - G_t(\mathbf{z})||_2 \, f(\mathbf{x}) \, d\mathbf{x}$$

$$+ \int_{\mathcal{X} \setminus (B \cup \mathcal{X}_\nu)} \min_{\mathbf{z}} ||\mathbf{x} - G_t(\mathbf{z})||_2 \, f(\mathbf{x}) \, d\mathbf{x} \quad (7)$$

$$\leq \int_B \min_{\mathbf{z}} ||\mathbf{x} - G_t(\mathbf{z})||_2 \, d\mathbf{x}$$

$$+ \nu \int_{\mathcal{X}_\nu} \min_{\mathbf{z}} ||\mathbf{x} - G_t(\mathbf{z})||_2 \, d\mathbf{x} \quad (8)$$

$$\leq \mu(B) \sup_{\mathbf{x} \in B} \min_{\mathbf{z}} ||\mathbf{x} - G_t(\mathbf{z})||_2$$

$$+ \nu\mu(\mathcal{X}_\nu) \sup_{\mathbf{x} \in \mathcal{X}_\nu} \min_{\mathbf{z}} ||\mathbf{x} - G_t(\mathbf{z})||_2 \quad (9)$$

$$\leq (\mu(B) + \nu\mu(\mathcal{X}_\nu)) \sup_{\mathbf{x} \in \mathcal{X}} \min_{\mathbf{z}} ||\mathbf{x} - G_t(\mathbf{z})||_2 \quad (10)$$

$$= (\mu(B) + \nu\mu(\mathcal{X}_\nu)) \max_{\mathbf{x} \in \mathcal{X}} \min_{\mathbf{z}} ||\mathbf{x} - G_t(\mathbf{z})||_2 \quad (11)$$

$$\leq C(\epsilon + \nu\mu(\mathcal{X})) \quad (12)$$

where $C > 0$ is a positive constant.

Equation (8) follows from the fact that $\min_{\mathbf{z}} ||\mathbf{x} - G_t(\mathbf{z})||_2 = 0$ for $\mathbf{x} \in \mathcal{X} \setminus (B \cup \mathcal{X}_\nu)$, $f(\mathbf{x}) \leq 1$ for $\mathbf{x} \in \mathcal{X}$, and $f(\mathbf{x}) \leq \nu$ for $\mathbf{x} \in \mathcal{X}_\nu$. Equation (11) is obtained using the extreme value theorem since $\mathcal{X}$ is compact. To prove (12) we proceed as follows:

$$\max_{\mathbf{x} \in \mathcal{X}} \min_{\mathbf{z}} ||\mathbf{x} - G_t(\mathbf{z})||_2 \leq \min_{\mathbf{z}} \max_{\mathbf{x} \in \mathcal{X}} ||\mathbf{x} - G_t(\mathbf{z})||_2 \quad (13)$$

$$\leq \max_{\mathbf{x} \in \mathcal{X}} ||\mathbf{x} - G_t(\bar{\mathbf{z}})||_2 \quad (14)$$

$$= C \quad (15)$$

Equation (13) follows from the max-min inequality. In (14), $\bar{\mathbf{z}}$ is such that $G_t(\bar{\mathbf{z}}) \in \mathcal{X}$. Such a $\bar{\mathbf{z}}$ always exists for $t \geq t_0$. Equation (15) follows from the fact that $\mathcal{X}$ is compact.

Therefore, we obtain (12) by noting that $\sup_{t \geq t_0} \max_{\mathbf{x} \in \mathcal{X}} \min_{\mathbf{z}} ||\mathbf{x} - G_t(\mathbf{z})||_2 \leq C$. Since $\mu(\mathcal{X})$ is a finite positive constant and (12) is satisfied for any $\epsilon, \nu > 0$, this proves the theorem. ∎

This theorem shows that the right-hand side of (4) is actually small, which justifies the setup adopted in [Bora et al., 2017]. However, the conditions for Theorem 1 may be too strict in practice. For example, [Goodfellow et al., 2014] assume that at every step of adversarial training, the discriminator $D$ is allowed to reach its optimal value given $G$, which might be numerically infeasible. Therefore, the convergence of $||G(\hat{\mathbf{z}}) - \mathbf{x}||_2$ might not be computationally attainable. To this end, we consider a task-aware GAN training, which allows $G$ to be optimized specifically for the task of reconstructing compressed measurements.

### Task-Aware GAN Training

To make the GAN training task-aware, we propose to jointly optimize $\mathbf{z}$ and train the GAN using these $\mathbf{z}$'s. This is outlined in Algorithm 1, which alternates between three optimizations on $\mathbf{z}$, $G$, and $D$, respectively. In particular, we add the GD steps in lines 5-7 of the algorithm to the original GAN training framework. This enables the discriminator and generator to be optimized on values of $\mathbf{z}$ which resemble the ones seen at test time. As previously mentioned, the original GAN training algorithm uses randomly generated $\mathbf{z}$ values to train $G$ and $D$. However, in our setting, the trained GAN will not be given random $\mathbf{z}$ values at test time, but rather specific $\mathbf{z}$'s selected to minimize a loss function. It is therefore beneficial to train the GAN on $\mathbf{z}$'s obtained through the same process. We note that the extra term $\lambda_{\text{prior}}||\mathbf{z}^{(i)}||_2^2$ in (16) comes from the negative log-likelihood of the Gaussian prior on $\mathbf{z}$ [Bora et al., 2017].

### GAN Training on Compressed Inputs

As mentioned earlier, a large set of non-compressed training data may not be available in practice. We, therefore, assume that a small (or empty) set of non-compressed training

**Algorithm 1** Task-aware GAN training algorithm.

1: **for** number of training iterations **do**
2:    Sample a batch of $s$ training examples $\{\mathbf{x}^{(1)}, \ldots, \mathbf{x}^{(s)}\}$.
3:    For all $i$, compute $\mathbf{y}^{(i)} = \mathbf{A}\mathbf{x}^{(i)} + \boldsymbol{\zeta}^{(i)}$.
4:    Initialize $s$ random latent variables $\{\mathbf{z}^{(1)}, \ldots, \mathbf{z}^{(s)}\}$ using a zero-mean Gaussian prior.
5:    Initialize $D$ and $G$.
6:    **for** $L$ steps **do**
7:      For all $i$, update $\mathbf{z}^{(i)}$ by GD on the loss:
$$||\mathbf{y}^{(i)} - \mathbf{A}G(\mathbf{z}^{(i)})||_2^2 + \lambda_{\text{prior}}||\mathbf{z}^{(i)}||_2^2 \quad (16)$$
8:    **end for**
9:    Update the discriminator by GD on the loss:
$$-\frac{1}{s}\sum_{i=1}^{s}[\log D(\mathbf{x}^{(i)}) + \log(1 - D(G(\mathbf{z}^{(i)})))] \quad (17)$$
10:   Update the generator by GD on the loss:
$$\frac{1}{s}\sum_{i=1}^{s}\log(1 - D(G(\mathbf{z}^{(i)}))) \quad (18)$$
11: **end for**
    **return** $\{\hat{\mathbf{z}}^{(1)}, \hat{\mathbf{z}}^{(2)}, \ldots\}, \hat{G}, \hat{D}$

---

**Algorithm 2** GAN training algorithm using compressed training data.

1: **for** number of training iterations **do**
2:    Sample a batch of $s_1$ non-compressed training examples $\{\mathbf{x}^{(1)}, \ldots, \mathbf{x}^{(s_1)}\}$.
3:    For all $i$, compute $\mathbf{y}^{(i)} = \mathbf{A}\mathbf{x}^{(i)} + \boldsymbol{\zeta}^{(i)}$.
4:    Sample a batch of $s_2$ compressed training examples $\{\tilde{\mathbf{y}}^{(1)}, \ldots, \tilde{\mathbf{y}}^{(s_2)}\}$.
5:    Initialize $s_1$ random variables $\{\mathbf{z}^{(1)}, \ldots, \mathbf{z}^{(s_1)}\}$ and $s_2$ random variables $\{\tilde{\mathbf{z}}^{(1)}, \ldots, \tilde{\mathbf{z}}^{(s_2)}\}$ using a zero-mean Gaussian prior.
6:    **for** $L$ steps **do**
7:      For all $i$, update $\mathbf{z}^{(i)}$ by GD on the loss:
$$||\mathbf{y}^{(i)} - \mathbf{A}G(\mathbf{z}^{(i)})||_2^2 + \lambda_{\text{prior}}||\mathbf{z}^{(i)}||_2^2 \quad (19)$$
8:      For all $i$, update $\tilde{\mathbf{z}}^{(i)}$ by GD on the loss:
$$||\tilde{\mathbf{y}}^{(i)} - \mathbf{A}G(\tilde{\mathbf{z}}^{(i)})||_2^2 + \lambda_{\text{prior}}||\tilde{\mathbf{z}}^{(i)}||_2^2 \quad (20)$$
9:    **end for**
10:   Update the discriminators by GD on the losses:
$$\frac{-1}{s_1}\sum_{i=1}^{s_1}\log D_1(\mathbf{x}^{(i)}) + \log(1 - D_1(G(\mathbf{z}^{(i)}))) \quad (21)$$
$$\frac{-1}{s_2}\sum_{i=1}^{s_2}\log D_2(\tilde{\mathbf{y}}^{(i)}) + \log(1 - D_2(\mathbf{A}G(\tilde{\mathbf{z}}^{(i)}))) \quad (22)$$
11:   Update the generator by GD on the loss:
$$\frac{1}{s_1}\sum_{i=1}^{s_1}\log(1 - D_1(G(\mathbf{z}^{(i)})))$$
$$+ \frac{1}{s_2}\sum_{i=1}^{s_2}\log(1 - D_2(\mathbf{A}G(\tilde{\mathbf{z}}^{(i)}))) \quad (23)$$
12: **end for**
    **return** $\{\hat{\mathbf{z}}^{(1)}, \hat{\mathbf{z}}^{(2)}, \ldots\}, \hat{G}, \hat{D}_1, \hat{D}_2$

---

data and a larger set of compressed training measurements are available. We modify the training algorithm to reflect this change. In particular, we train two discriminators and a single generator using a combination of compressed and non-compressed training data. The first discriminator is used to distinguish between actual training data $\mathbf{x}$ and generated data $G(\mathbf{z})$. The second discriminator is used to distinguish between actual compressed training data $\mathbf{y}$ and generated data $\mathbf{A}G(\mathbf{z})$. The details of the training procedure can be found in Algorithm 2. The addition of a second discriminator in Algorithm 2 does not affect the representative power of the generator. In fact, similar arguments as in [Goodfellow et al., 2014, Proposition 2] can be made to show that, with the two discriminators, the distribution of the generator output being the same as that of the training data remains optimal.

### Contrastive Loss Regularization for Supervised Learning Tasks

The low-dimensional vector $\hat{\mathbf{z}}$, returned by Algorithm 1 or 2, can be used as an input to $\hat{G}$ to recover the original image $\mathbf{x}$. Since $\hat{G}$ learns to represent the overall data distribution of $\mathbf{x}$, $\hat{\mathbf{z}}$ must hold characteristic information specific to $\mathbf{x}$. This motivates us to use $\hat{\mathbf{z}}$ as an input feature for inference tasks such as classification. Since $\hat{\mathbf{z}}$ is of much lower dimension than $\mathbf{x}$ (and, usually, $\mathbf{y}$), using it as an input feature to a classifier reduces the curse of dimensionality. To drive $\hat{\mathbf{z}}$ to be more discriminative for the classification task, we add a contrastive loss [Chopra, Hadsell, and LeCun, 2005] term to (16). We assume that labeled training data is available, and the ground-truth label of $\mathbf{x}^{(i)}$ is denoted by $\ell_i$. The contrastive loss of a batch of $\mathbf{z}$'s is given by:

$$\mathcal{L}_{\text{contr}} \triangleq \frac{\lambda_{\text{contr}}}{s(s-1)} \sum_{i,j=1}^{s} \left[ \mathbb{1}(\ell_i = \ell_j)||\mathbf{z}^{(i)} - \mathbf{z}^{(j)}||_2^2 \right.$$
$$\left. + \mathbb{1}(\ell_i \neq \ell_j)\max\{0, M - ||\mathbf{z}^{(i)} - \mathbf{z}^{(j)}||_2^2\} \right] \quad (24)$$

where $\lambda_{\text{contr}}$ is a weight which dictates the relative importance of this loss, and $M$ is a positive margin.

### Experiments

In our experiments, we use three different image datasets: the MNIST handwritten digits dataset [LeCun et al., 1998], the Fashion-MNIST (F-MNIST) clothing articles dataset [Xiao, Rasul, and Vollgraf, 2017], and the CelebFaces Attributes dataset (CelebA) [Liu et al., 2015].

The MNIST and F-MNIST datasets each consists of 60,000 training images and 10,000 testing images, each of size $28 \times 28$. We split the training images into a training set of 50,000 images and hold-out a validation set containing 10,000 images. The testing set is kept the same. The images contain a single channel, therefore the input dimension $n$ is $28 \times 28 = 784$.

The CelebA dataset is a large-scale face dataset consisting of more than 200,000 face images, split into training, validation, and testing sets. The RGB images were cropped to a size of $64 \times 64$, resulting in an input dimension of $n = 64 \times 64 \times 3 = 12,288$.

For all datasets, our generative and discriminative models follow the Deep Convolutional GAN (DCGAN) architecture in [Radford, Metz, and Chintala, 2016]. We use the Adam optimizer [Kingma and Ba, 2014] for training the GAN. All hyper-parameters were either set to match the ones in [Bora et al., 2017] or chosen by testing on the holdout validation set. Our implementation is based on TensorFlow and builds on open-source software [Kim, 2017, Bora et al., 2017]. Details of the hyper-parameters used in our experiments can be found in the code repository.

## Reconstruction

We first perform a compressed sensing reconstruction task. We train our model using Algorithm 1, assuming access to the original non-compressed training set. We refer to our trained model as Compressed Sensing GAN (CSGAN), since the GAN was trained in a task-aware fashion for compressed sensing. As a baseline, we compare our reconstruction results to those obtained by the method in [Bora et al., 2017], which trains a DCGAN using the usual GAN training framework. At test time, both methods optimize (16) to obtain $\hat{\mathbf{z}}$, with the same learning rate and number of GD iterations. For both cases, we perform the same number of random restarts on the initialization of $\mathbf{z}$. The reconstruction is then given by $G(\hat{\mathbf{z}})$.

Additionally, we compare the results to Lasso, performed directly on the pixel values for MNIST and F-MNIST, and in the Discrete Cosine Transform (DCT) and Wavelet Transform domains for CelebA as was done in [Bora et al., 2017]. We also compare our results to two iterative shrinkage-thresholding algorithms: the Two-step Iterative Shrinkage-Thresholding algorithm (TwIST) [Bioucas-Dias and Figueiredo, 2007] and the Fast Iterative Shrinkage-Thresholding algorithm (FISTA) [Beck and Teboulle, 2009] and, in the case of MNIST and F-MNIST, to reconstructions based on the Split Bregman (SB) method with a total variation (TV) regularizer [Goldstein and Osher, 2009], and the SB method with a Besov norm regularizer [Yin et al., 2008]. The SB method was not performed on the CelebA dataset as the smoothness assumption is not applicable in the case of RGB images when different channels are not treated independently. We report per-pixel mean-squared reconstruction error results in Figure 1, as we vary the number of measurements $m$. It is shown that, and especially for very low values of $m$, the task-aware training of CSGAN is able to more reliably reconstruct unseen samples $\mathbf{x}$.

**Remark 1** *We note that the DCGAN results for MNIST in Figure 1 differ from those reported in [Bora et al., 2017], due to the use of a GAN instead of a VAE. As GANs and VAEs vary in their training methods, for clarity of presentation, we have opted to only use GANs in this paper. However, our method can be readily extended to VAE models.*

### GAN Training on Compressed Inputs

As previously mentioned, for some applications, it might be prohibitive to acquire a large training set consisting of non-compressed images. However, compressed training data can be readily available. To empirically validate the dual discriminator training method on compressed measurements and non-compressed inputs, we study the effect of varying the size of the non-compressed training set. Naturally, DCGAN can only be trained on the non-compressed training images, and suffers from over-fitting. The results are reported in Tables 1 and 2, and Figure 2, where NC = Number of non-compressed training samples. We can see that the addition of a compressed data discriminator has successfully allowed the training of a CSGAN using compressed measurements. Additionally, we note an interesting trend: when NC $= 0$, CSGAN performs better than when NC $= 100$ and $1,000$ (but not when NC $= 8,000$). In fact, when the discriminator for non-compressed data $D_1$ overfits the small amount of training data, this negatively affects the performance of the generator (which is shared by both discriminators $D_1$ and $D_2$). In such cases, it is beneficial to only use the compressed data discriminator $D_2$. The smallest number of non-compressed data needed to train $D_1$ can be determined using the validation set. Generally, we can see that the compressed data discriminator is extremely useful especially when the amount of available non-compressed training data is very low.

| NC | DCGAN | CSGAN |
|---|---|---|
| 0 | - | **0.0299** |
| 100 | 0.1138 | **0.1053** |
| 1,000 | 0.0859 | **0.0322** |
| 8,000 | 0.0894 | **0.0124** |

Table 1: MNIST: Reconstruction results for $m = 200$ when varying the number of non-compressed training data.

| NC | **A** random Gaussian | | Super-resolution | |
|---|---|---|---|---|
|  | DCGAN | CSGAN | DCGAN | CSGAN |
| 1,000 | 0.1278 | **0.0514** | 0.1006 | **0.0510** |
| 4,000 | 0.0837 | **0.0394** | 0.0582 | **0.0436** |
| 32,000 | 0.0800 | **0.0308** | **0.0241** | 0.0247 |

Table 2: CelebA: Reconstruction results for $m = 500$ when varying the number of non-compressed training data.

In the extreme case, where only compressed measurements are available for training, we show qualitative results of MNIST and F-MNIST reconstruction in Figures 2 and 3. We would like to emphasize that this CSGAN has never seen any real training image and has been solely trained on compressed measurements, yet can reconstruct reasonably good samples. Additionally, quantitative reconstruction results for various values of the number of measurements $m$ can be found in Table 3.

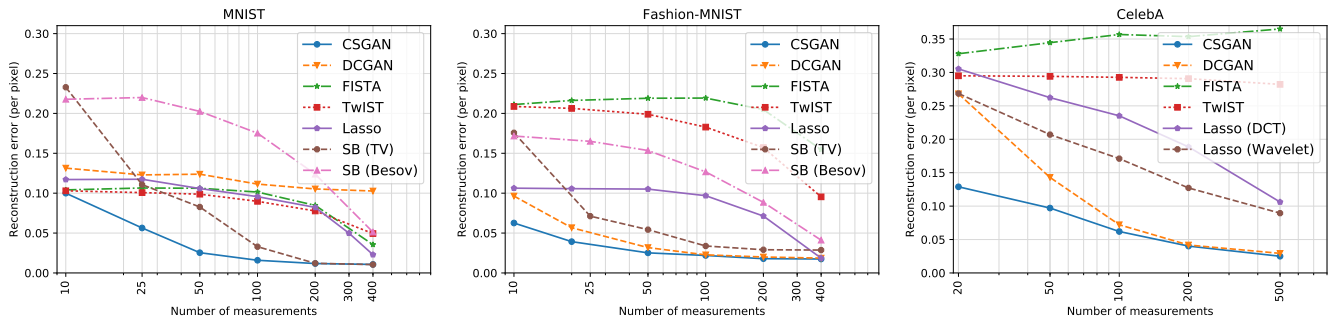

Figure 1: MNIST, F-MNIST, and CelebA reconstruction results for various measurements numbers $m$.

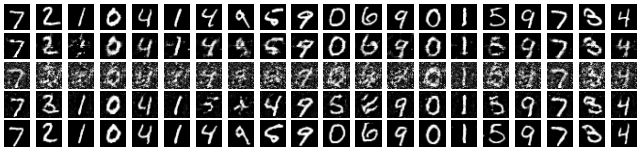

Figure 2: MNIST reconstruction results with $m = 200$. Top to bottom rows: original images, reconstructions with NC = 0, reconstructions with NC = 100, reconstructions with NC = 1,000, and reconstructions with NC = 8,000.

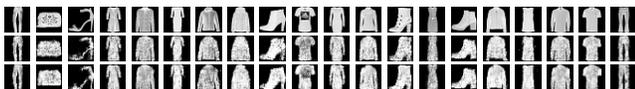

Figure 3: F-MNIST reconstruction results when only compressed training data is available. Top row: original image; middle row: reconstructed image from $m = 200$ measurements; bottom row: reconstructed image from $m = 400$ measurements.

| $m$ | MNIST | F-MNIST |
|---|---|---|
| 20 | 0.2164 | 0.2829 |
| 50 | 0.0535 | 0.0988 |
| 100 | 0.0304 | 0.0534 |
| 200 | 0.0299 | 0.0579 |

Table 3: CSGAN reconstruction results when only compressed training data is available (NC = 0) for various measurements numbers $m$.

## Super-Resolution

Super-resolution is the task of increasing the resolution of an image. For this special case, where **A** is a matrix that averages neighboring pixels, no theoretical guarantees (such as (4)) are known. However, experiments using such averaging matrices **A**'s still provide good results. Super-resolution is actually a relevant application where an abundance of non-compressed (i.e., high resolution) images may not be available. Results when varying the number of non-compressed training data are reported in Table 2 for compressed measurements four times smaller than the original images ($3,072$ measurements). Additionally, qualitative results can be seen in Figure 4 on the CelebA dataset. We can see that CSGAN produces realistic reconstructions that resemble the original image.

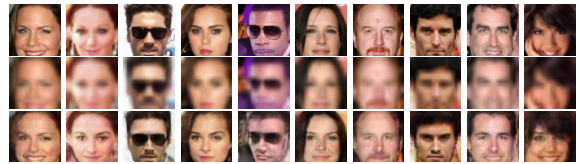

Figure 4: CelebA super-resolution results. Top row: original image; middle row: blurred image; bottom row: reconstructed image.

## Classification

In this section, we use the discriminatively-regularized CSGAN with the additional contrastive loss defined in (24), with $\lambda_{contr} = 100$ and $M = 0.1$. We train a Convolutional Neural Network (CNN) classifier based on the LeNet [LeCun et al., 1998] architecture, with a fully-connected layer to map the input latent variables $\hat{z}$ to a $784$-dimensional vector as expected by LeNet. We train the network over 30 epochs and pick the best model based on the holdout validation set. Our results are reported in Tables 4 and 5. We can see that inference can indeed be made even using an extremely small number of measurements. When training this CNN using the $\hat{z}$'s obtained from DCGAN, we obtain much lower classification accuracies. This clearly demonstrates the effectiveness of the regularization of CSGAN using the contrastive loss.

In order to further investigate the structure of the GAN latent space, we report classification accuracies using a basic 50-nearest-neighbor (50-NN) classifier based on the Eu-clidean distance in Tables 5 and 6. This simple 50-NN classifier clearly does not give state-of-the-art classification performance. It, however, serves to show that the latent space has indeed been regularized so that samples belonging to the same class are represented by $\mathbf{z}$'s which are close to each other in the Euclidean distance sense. This is made even clearer when compared to the performance of the same 50-NN classifier on the DCGAN latent space.

Finally, we report per-pixel mean-squared reconstruction error results on the MNIST and F-MNIST datasets when using the contrastive loss regularizer in Table 7. These results serve to show that the addition of the discriminative regularizer does not hurt reconstruction performance.

| $m$ | SF | LeNet | |
| --- | --- | --- | --- |
| | | CSGAN + cont. $\hat{\mathbf{z}}$ | DCGAN |
| 8 | 0.3697 | **0.4560** | 0.3814 |
| 39 | 0.4679 | **0.7572** | 0.4304 |
| 78 | 0.5645 | **0.8740** | 0.4296 |
| 196 | 0.7258 | **0.9257** | 0.4818 |

Table 4: Classification accuracy on MNIST using Smash Filters (SF) [Davenport et al., 2007] and the LeNet CNN classifier.

| $m$ | LeNet | | 50-NN | |
| --- | --- | --- | --- | --- |
| | CSGAN + cont. $\hat{\mathbf{z}}$ | DCGAN | CSGAN + cont. $\hat{\mathbf{z}}$ | DCGAN |
| 10 | **0.4881** | 0.4372 | **0.3937** | 0.3019 |
| 50 | **0.7410** | 0.6780 | **0.6073** | 0.4183 |
| 100 | **0.7705** | 0.7363 | **0.6377** | 0.4495 |
| 200 | **0.7830** | 0.7584 | **0.6456** | 0.4522 |

Table 5: Classification accuracy on F-MNIST using the LeNet CNN and 50-NN classifiers.

| $m$ | CSGAN + cont. $\hat{\mathbf{z}}$ | DCGAN |
| --- | --- | --- |
| 8 | 0.3561 | **0.3679** |
| 39 | **0.5987** | 0.3951 |
| 78 | **0.6991** | 0.3957 |
| 196 | **0.7656** | 0.4555 |

Table 6: Classification accuracy on MNIST using a 50-NN classifier.

## Conclusion

In this paper, we present an effective method for training task-aware generative models, specifically for compressive sensing tasks. We show that this task awareness improves the performance, especially when a very low number of measurements is available. Additionally, we demonstrate that it is also possible to train CSGANs with only compressed measurements as training data, or, if available, only a small number of non-compressed measurements. In the future, we would like to train the CSGAN and classifier jointly in an end-to-end manner.

| $m$ | MNIST | | F-MNIST | |
| --- | --- | --- | --- | --- |
| | CSGAN | CSGAN + cont. $\hat{\mathbf{z}}$ | CSGAN | CSGAN + cont. $\hat{\mathbf{z}}$ |
| 10 | 0.1042 | **0.0999** | **0.0627** | 0.0732 |
| 50 | 0.0353 | **0.0334** | **0.0253** | 0.0254 |
| 100 | 0.0285 | **0.0186** | 0.0220 | **0.0203** |
| 200 | 0.0199 | **0.0139** | 0.0179 | **0.0179** |
| 400 | 0.0169 | **0.0112** | 0.0175 | **0.0168** |

Table 7: Per-pixel mean-squared reconstruction error results when using the contrastive loss regularizer (with $\mathbf{z}$ dimension $k = 20$).

**Acknowledgment** This research is based upon work supported by the Office of the Director of National Intelligence (ODNI), Intelligence Advanced Research Projects Activity (IARPA), via IARPA R&D Contract No. 2014-14071600012. The views and conclusions contained herein are those of the authors and should not be interpreted as necessarily representing the official policies or endorsements, either expressed or implied, of the ODNI, IARPA, or the U.S. Government. The U.S. Government is authorized to reproduce and distribute reprints for Governmental purposes notwithstanding any copyright annotation thereon.